\documentclass[a4paper,twoside]{article}

\usepackage{epsfig}
\usepackage{subcaption}
\usepackage{calc}
\usepackage{amssymb}
\usepackage{amstext}
\usepackage{amsmath}
\usepackage{amsthm}
\usepackage{multicol}
\usepackage{pslatex}
\usepackage{apalike}
\usepackage{SCITEPRESS}     % Please add other packages that you may need BEFORE the SCITEPRESS.sty package.
\usepackage[table]{xcolor}
\usepackage{booktabs}

\usepackage[colorlinks=true,citecolor=blue]{hyperref}
\usepackage[edges]{forest}
\tikzset{%
    parent/.style =          {align=center,text width=3.5cm,rounded corners=3pt},
    child/.style =           {align=center,text width=2.5cm,rounded corners=3pt},
    grandchild/.style =      {align=center,text width=2cm,rounded corners=3pt},
    greatgrandchild/.style = {align=center,text width=1.5cm,rounded corners=3pt},
    referenceblock/.style =  {align=center,text width=1.5cm,rounded corners=2pt}
}

\begin{document}

\title{Simulation-to-Reality domain adaptation for offline 3D object annotation on pointclouds with correlation alignment}

% \author{
% \authorname{Anonymous}
% \affiliation{Anonymous}
% }
\author{\authorname{Weishuang Zhang, B Ravi Kiran, Thomas Gauthier, Yanis Mazouz, Theo Steger}
\affiliation{Navya, France}
\email{\{first\_name.second\_name\}@navya.tech}
}


\abstract{
   Annotating objects with 3D bounding boxes in LiDAR pointclouds is a costly human driven process in an autonomous driving perception system. In this paper, we present a method to semi-automatically annotate real-world pointclouds collected by deployment vehicles using simulated data. We train a 3D object detector model on labeled simulated data from CARLA jointly with real world pointclouds from our target vehicle. The supervised object detection loss is augmented with a CORAL loss term to reduce the distance between labeled simulated and unlabeled real pointcloud feature representations. The goal here is to learn  representations that are invariant to simulated (labeled) and real-world (unlabeled) target domains. We also provide an updated survey on domain adaptation methods for pointclouds.
}

\onecolumn \maketitle \normalsize \setcounter{footnote}{0} \vfill

%%%%%%%%% BODY TEXT
\section{Introduction}

Many self-driving vehicles (SDV) rely on LiDAR (Light Detection And Ranging) technology to perceive their surroundings. There are multiple real-world SDV largescale LiDAR annotated datasets including KITTI \cite{geiger2013vision}, nuScenes \cite{caesar2019nuscenes}, Waymo \cite{sun2020scalability}, Lyft \cite{lyft}, Semantic-KITTI \cite{behley2019semantickitti}, nuScenes LiDAR-Seg \cite{fong2021panoptic}. 
This has provided a large performance gain across various supervised 3D detection and segmentation perception pipelines. Though generating annotated pointcloud datasets is a costly, meticulous \& time consuming process requiring a large number of human annotators. Annotating real data also poses physical constraints on the position, number of obstacles as well as type of dynamic scenarios. 

Simulators have become a cheaper and scalable alternative in terms of scenario diversity and time of training. In comparison to simulation, real world annotation pipelines have these key operational issues: 
\begin{enumerate}
    \item Ensuring sensor calibration and synchronization (e.g. Camera-Lidar or Radar-Lidar) to achieve precise annotations when the pointclouds are sparse.
    \item Most road-datasets contain car as majority classes due to the domain of operation.  Though in real world operations, certain zones can contain more pedestrians. There is a change in class distribution between the training and test domains.
    \item Furthermore, annotation is often performed on dense LiDAR pointclouds (64/32 layers). Transfer learning from datasets to sparse multi-Lidar pointclouds remains a big challenge.
\end{enumerate}

To address these problems, autonomous driving simulators such as CARLA \cite{Dosovitskiy17CARLA} can  provide inexpensive source of synthetic annotated data. Our contributions include :
\begin{itemize}
    \item A short incremental review on the state of domain adaptation for pointclouds tasks, classification, semantic segmentation \& detection.
    \item A case study on 3D-object detection on our deployment vehicle's pointclouds which evaluates the application of domain invariant representation learning using the correlation alignment loss (CORAL) between simulation and real pointclouds
    \item A qualitative analysis of the sources of the domain gap between simulated and real pointclouds.
\end{itemize}

\subsection{Domain adaptation (DA) on pointclouds} 

\begin{table*}[ht]
\centering
 \begin{tabular}{p{2.5cm} p{12.5cm}} \toprule
 DA  Methods & Description\\ \midrule 
 Aligning I/O representations & Dataset-to-Dataset(D2D) transfer \cite{triess2021ivSurveyDA} involves transfer learning between LiDAR Datasets collected with different LiDAR configurations (number, scanning patter, spatial resolution, different classes/label spaces) would require alignment either by upsampling, downsampling, re-sampling of pointclouds.  \cite{icinco21} class sensitive data
augmentation. These methods are frequently hand-engineered. \cite{tomasello2019dscnet} present a Deep Sensor Cloning methods which enables the generation of pointclouds from expensive LiDARs (HDL64) using CNNs along with in-expensive LiDARs (Scala). \\ 
 \hline
Modeling Physics & Authors \cite{hahner2021fog} have proposed a fog simulation method in pointclouds that is applicable to any
LiDAR dataset. \cite{zhao2020epointda} learn dropout noise from real world data.\\
 \hline
 Adversarial Domain Mapping (ADM) &  Learns a conditional mapping from source domain samples to their target domain samples using Generative Adversarial Networks (GANs).  ADM can enable Simulation-to-Real(S2R) DA. Annotations from simulation can be leveraged by mapping Simulated clouds to real target domain clouds with subsequent training using source domain labels. Authors \cite{sallab2019lidar} map simulated BEV images to real world equivalents while improving object detection performance. \\ 
 \hline
 Domain Invariant Learning & These methods are usually adversarial methods that align the feature spaces between source \& target pointcloud domains, thus enforcing consistent prediction on target domain.  CORAL loss based on \cite{sun2017correlation} belongs to this family. Authors \cite{langer2020domain} generate semi-synthetic pointclouds from the source data  while performing correlation alignment between synthetic target scans and target scans.\\
 \hline
 Simulation-To-Real (S2R) & These families of methods focus on reducing the domain gap between simulation and reality. Authors \cite{debortoli2021adversarial} claims to have encouraged the 3D feature encoder to extract features that
are invariant across simulated and real scenes. Authors \cite{Huang2021GenerationFA}  generate synthetic pointclouds to
train classification models instead of aligning features. They are thus able to highlight which part of the object is
transferred. \\
   \bottomrule
 \end{tabular}
 \vspace{5pt}
 \caption{Categorization of Domain adaptation methods for LiDAR pointclouds.}
 \label{tab:DA_review}
\end{table*}

In a typical deep learning application such as object detection using LiDAR or Camera, a crucial assumption made is that the training dataset domain (also called source domain) and test data domain (target domain) share the same feature space distribution. This could be broken in multiple ways (non IID sampling, IID referring to Independent and identically distributed) and is a key constraint in the performance of DNNs in open operational domains. Domain adaption is a set of transformations (or representation learning) that aligns the features between source and target domains. Based on the availability of labels in the target domain, DA can be supervised or unsupervised. Transfer learning is a subset of supervised DA where labels in source domain can be used to fine tune DNNs to their target domains, though this is usually a costly process. Unsupervised DA usually operates in target domains where there are either no or very few labels.

Authors \cite{triess2021ivSurveyDA} and \cite{besic2021unsupervised} provide surveys on DA methods for the perception tasks (segmentation/detection/classification of pointclouds) in LiDAR. We provide a summary of the taxonomy of methods described by this survey in table \ref{tab:DA_review}. We have updated the survey with new references and methods from recent literature. Majority of these methods are focused on unsupervised DA where there are no target domain labels available. The goal here is to highlight potential DA methods that could be used to perform Simulation-To-Real(S2R) domain adaptation.

\begin{figure*}[ht!]
\centering
    \includegraphics[width =0.8\textwidth]
    {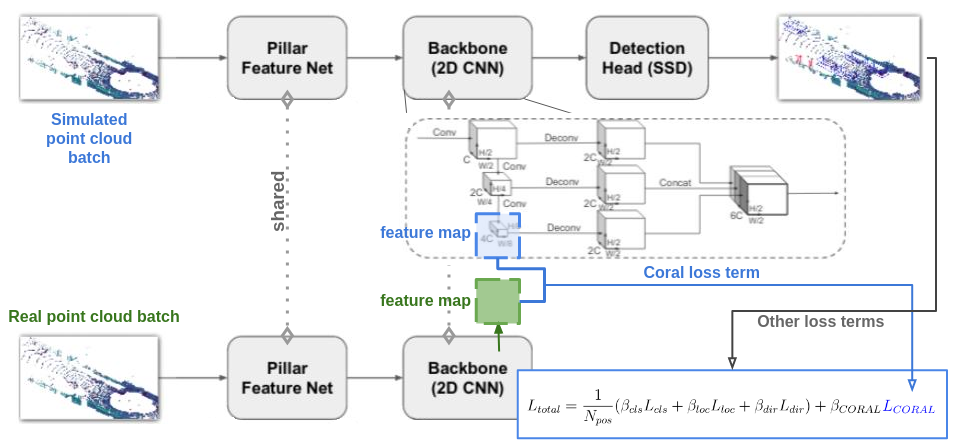}
    \caption{We reproduce the Pointpillars pipeline along with the CORAL loss between real \& simulated feature maps. }
    \label{fig:CORAL_loss_in_SECOND}
\end{figure*}

\subsection{Simulation-To-Real DA}  
In this subsection, we summarize key studies using simulation-to-real(S2R) domain adaptation (DA) methods. This implies pre-training on simulated pointclouds while evaluating on real-world pointclouds. Authors \cite{yue2018lidar} demonstrate the first pointcloud simulator while showing a significant improvement in accuracy (+9\%) in pointcloud semantic segmentation by augmenting the training dataset with the generated synthesized data. 
% \cite{tomasello2019dscnet} present a Deep Sensor Cloning (DSC) network that is able to generate the range and return outputs of expensive (VLP64-LiDAR) and in-expensive LiDAR sensors setups (Ibeo Scala LIDAR). 
Another key issue in simulating LiDAR pointclouds is generating sensor \& material dependant intensity channel output. Most simulators do not model the intensity function, though mostly modeling ray tracing and illumination operations. Authors \cite{wu2019squeezesegv2} proposed a learned intensity rendering network which regresses the intensity as a function of the xyz coordinate channels in the range image. However this mapping from xyz-intensity is a highly non-stationary function, since similar geometrical surfaces (such as building wall and metal) could have drastically different intensity values. Thus learning these mappings is a difficult process. \cite{vacek2021learning} propose the use of RGB data along with xyz coordinates to improve intensity regression on polished parts of car bodyworks, windows, traffic signs and license/registration plates.

\cite{carlaSimObjectDet2019} demonstrate the Sim-to-Real transferability for 3D object detection between CARLA and KITTI, using different mixtures (training on different combinations of datasets, sequential fine tuning on the 2 datasets) of real and simulated data to train Yolo3D, Voxelnet and Pointpillar architectures. They demonstrate there are significant gains in performance for object detection. \cite{brekke2019multimodal} evaluate simulated pre-training on both camera images and lidar scans from CARLA, while training a AVOD-FPN network. Authors remark that real world data cannot be replaced though simulated data can considerably reduce the amount of training data required to achieve target accuracy levels in the detection task.

Authors \cite{fang2020augmented} and \cite{manivasagam2020lidarsim}  perform Real-To-Simulation (R2S) modeling of pointclouds, where real-world scans (3D maps of background) are used to build a catalog of diverse scenarios. Recorded dynamic objects are then inserted into these existing scenarios. This is then used to train a robust 3D object detection. Authors \cite{saltori2020sf} study dataset-to-dataset domain adaptation between KITTI-nuScenes leveraging motion coherence across detections, reversible scale transformations and pseudo-annotations.
Authors \cite{deschaud2021pariscarla3d} have created the Paris-CARLA-3D dataset, with simulated pointcloud and camera data generated in CARLA while the real world data was logged in Paris. The goal of the dataset is to evaluate the unsupervised domain adaptation from CARLA to Paris data.

Authors \cite{meng2020weakly} have also explored weakly-supervised learning (annotating horizontal centers of objects in bird’s view scenes) while learning to predict a full 3D bounding box. 
\begin{figure*}[ht!]
    \centering
    \includegraphics[width=0.2\textwidth, height=4.5cm]{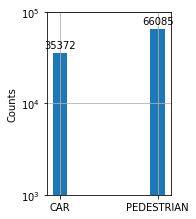}
    \hspace{1cm}
    \includegraphics[width=0.25\textwidth]{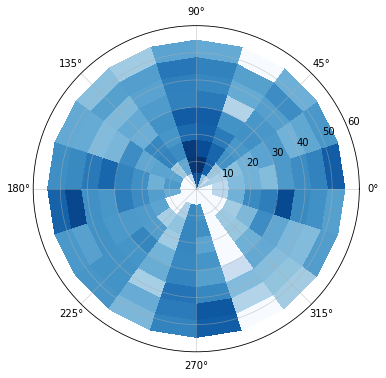}
    \hspace{1cm}
    \includegraphics[width=0.25\textwidth]{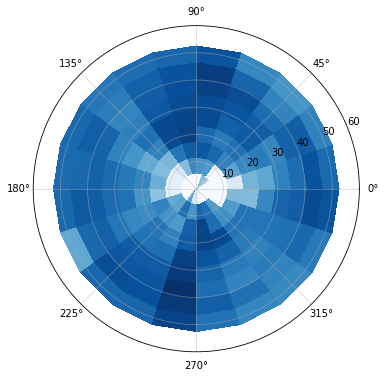}
    \caption{Class histogram in simulated dataset, along with polar density map for box annotations in dataset. The intensity values were log-scaled.}
    \label{fig:sim_histogram}
\end{figure*}

\begin{figure}
\begin{minipage}{0.45\textwidth}
    \centering
    \includegraphics[width=\textwidth]{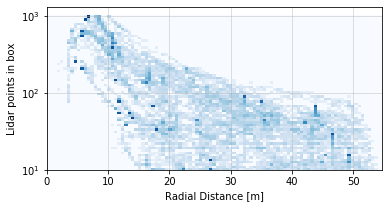}
        Car
\end{minipage}
\begin{minipage}{0.45\textwidth}
    \centering
    \includegraphics[width=\textwidth]{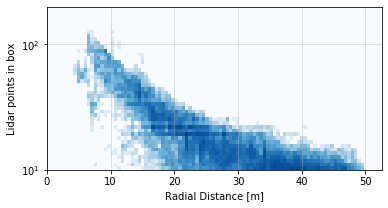}
        Pedestrian
\end{minipage}
\caption{Number of points with ground truth boxes as the range of the box varies in the simulated dataset.}
\label{fig:sa_anno_density}
\end{figure}

\section{Simulation-to-Real on Custom LiDAR dataset}
In this section we focus on our experimental demonstration.
Our goal here is to summarize the different components of the simulation-to-real domain adaptation experiment, namely the object detection architecture on pointclouds, the features being used by the CORAL loss to perform correlation alignment between feature maps. This experiment shall be carried out on our proprietary simulated and real pointcloud dataset.

\subsection{Simulated \& Real datasets}
The simulated pointcloud dataset was generated with 16k multi-lidar scans.
There was no "Cyclist" class included in the dataset and mainly constituted of "Pedestrian" and "Car" classes. 
The class frequencies in the simulated dataset is show in figure \ref{fig:sim_histogram}. The plots also demonstrate the polar histograms (range, azimuth), where each cell in the plot contains the normalized frequencies of bounding boxes visible across the dataset.

\paragraph{Real pointclouds} 
The real pointcloud dataset was constructed on board the target vehicle in operation at our test site.  

\textit{Labeled :} We annotated a small dataset of LiDAR pointclouds coming from the vehicles real domain of operation. This contains merely 223 scans, arranged in 4 continuous sequences. Each LIDAR scan contains 16.9 points, 2k in minimum and 20.7 in maximum, with a median equals to 16.7k.

\textit{Unlabeled:} To performed unsupervised domain adaptation we used a large collection (1000 scans) of LiDAR pointclouds coming from the vehicles real domain of operation with different obstacles (pedestrian and cars mainly) in varied configurations.

\paragraph{Point density per box} The plot in figure \ref{fig:sa_anno_density} shows the number points within ground truth boxes for each category in the y axis, while the range/radius at which the box is present. The plot for the car category is very interesting as it doesn't follow the pattern in the public dataset, the relationship between point density per box and distance to box center is looser in this dataset. This empirical parameter is key to ensure robust feature extraction for the car category at different point density within each bounding box. Filtering out bounding boxes that contain very few points in the real world dataset thus is demonstrated as a key manual engineering step that directly affects the quality of features being extracted within any given object detection framework.

\subsection{Pointpillars architecture}

The pipeline of the Pointpillars consists of three main parts: Pillar Feature Network, Backbone and SSD Detection Head, as shown below.

\textbf{Pillar Feature Net} is a feature encoder network that converts a pointcloud to a sparse pseudo-image composed by two modules. The input pointcloud is first discretised into an evenly spaced grid in the x-y plane. A tensor of size (D,N) is then calculated for each voxel, where D is the dimension of feature for each sampled point in the pillar and N is the maximum number of points per pillar. Thus the pointcloud is converted into a stacked-pillars tensor of size (D,P,N), where P denotes the number of non-empty pillars per pointcloud.

A layer consists of a Linear-BatchNorm-ReLU follows to extract pillar-wise features, with max pooling over the channels to create an output tensor of size (C,P). Then the features are scattered back to the original pillar locations to create a pseudo-image of size (C,H,W), where C is number of channels fixed to 64.

\textbf{Backbone} consists in blocks of top-down 2D CNNs, which can be characterized by a series of blocks. Each top block has 2D convolutional layers to reduce the 2D tensor sizes into half, followed by BatchNorm and a ReLU. The processed tensors are combined through upsampling and concatenation, as shown above.

\textbf{SSD Detection Head} is a detection head that finally detects and regresses 3D 
boxes in Pointpillars. The prior boxes are matched to the ground truth using IoU.

\begin{figure*}[ht]
\centering
    \begin{minipage}{1.0\textwidth}
        \includegraphics[height = 4.5cm]
        {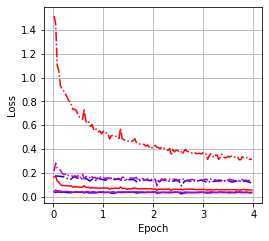}
        \includegraphics[height = 4.5cm]
        {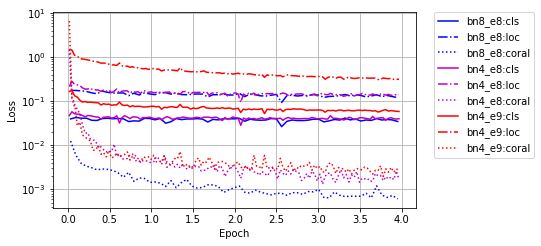}
        \caption{The plot on the left represents the global loss, while plot on the right represent the localization, classification and Coral Loss curves. Different experiments with batch sizes 4 \& 8 (bn4 and bn8 respectively), and CORAL loss weighting $\beta_{DA} = 1e8$ \& $1e9$ are demonstrated. The loc, cls refer to the localization and classification losses.}
        \label{fig:CORAL_loss-sim_real_asterix}
    \end{minipage}
\end{figure*}

\subsection{Coral Loss}
\label{chap:DomainAdaptation}

We use the CORAL loss \cite{sun2017correlation} to reduce the geodesic distance  between the simulated and real pointclouds. The same method could also be used to minimize distance between embeddings of pointclouds coming from different LiDAR configurations.

The CORAL loss is described between two domains for a single feature layer. Given 
source-domain training examples $D_{S} = \{x_{i}\}, x \in R^{d}$ with
labels $L_{s} = \{y_{i}\}, i \in\{1, 2, ..., L\}$, and unlabeled target data
$D_{T} = \{u_{i}\}, x \in R^{d}$. Suppose the number of source and target data are
${n_{S}}$ and ${n_{T}}$ respectively. Here, both $x$ and $u$ are the d-dimensional deep
layer activations $\phi(I)$ of input $I$ that we are trying to learn. 
% \tga{Should not we introduce extra notations for the activations ? x and u were the inputs, no ?}
Let $D_{S}^{ij}$/$D_{T}^{ij}$ be the j-th dimension of the i-th source/target data, $C_{S}$/$C_{T}$ denote the 2D feature covariance matrices. The CORAL loss ie. the distance between the source and target features is defined as:
\begin{equation}
    L_{DA}= \frac{1}{4d^{2}}||C_{S} - C_{T}||^{2}_{F}
\end{equation}
where $|| \cdot ||^{2}_{F}$ denotes the squared matrix Frobenius norm.

\subsection{Pipeline}

Figure \ref{fig:CORAL_loss_in_SECOND} shows the Pointpillars architecture using CORAL loss for deep domain adaptation. 

The two mini-batches batches of simulate and real pointclouds pass through the shared backbone pipeline. The CORAL loss is evaluated at the end of the 2D backbone as the feature map considering its reduced size as shown in figure \ref{fig:CORAL_loss_in_SECOND}. 

The shape of this feature map is (bn, 4C, H/8, W/8), where bn is the batch size, C is the channel number output by PFN, which is set to 64, while H and W represent the size of the pillar grids in the xy plane.

For a pointcloud ranged in $\pm50m$ both in x and y with a grid size of 0.25m, the feature map shape in our experiment is (bn, 256, 50, 50). As the CORAL loss $L_{CORAL}$ needs a pair of 2-dimension inputs, we chose to reshape the feature map into (256bn, 2500). And the Loss function is modified to:
\begin{equation}
L_{total}=\frac{1}{N_{pos}}(\beta_{cls}L_{cls}+\beta_{loc}L_{loc} +\beta_{dir}L_{dir}) + \beta_{DA}L_{DA}
\end{equation}
Where $\beta_{DA}$ is the weight for the CORAL loss. 
$L_{cls}, L_{loc}$ and $L_{DA}$ represent the classification, localization and CORAL domain adaptation loss terms.

\section{Experiment \& Results}
In this section, We describe the dataset setup and experiments performed with different hyperparameters to demonstrate the effect of adding a domain adaptation loss based on CORAL.

In our study we use the data from simulation which contains 12646 scans as labeled pipeline input, and 2 sequences of unlabeled real vehicle data which contains around 500 scans. Both of the input pointclouds are ranged within 50m in both x and y axis, $[-3, 1]$m in z axis. Similar to what is done in nuScenes, the point pillar is sized to 0.25m×0.25m with max number of points per pillar set to 60. Zero padding is applied for not fully filled pillars.

Performance metrics measured were using the official KITTI evaluation toolkit \cite{geiger2013vision} and nuScenes devkit. The four metrics used in our experiments denote respectively as mean average precision in BEV, 3D, 2D (image plane), and Average Orientation Similarity (AOS). In this section we'll step from basic conceptions to explain how these metrics are calculated. Some of them are not directly used but fundamental to understand the PASCAL criteria introduced in object detection. 

The performance metrics of simulation trained model on real data can be seen in table \ref{tab:sim2real} .
We present 3 experiments on sim-to-real domain adaptation,
the loss curves are shown below in \ref{fig:CORAL_loss-sim_real_asterix}. 
The three models are represented by different colors, where the blue ones denotes the model with batch size 8, CORAL loss weight $\beta_\text{DA} = 1e8$, the violet ones denote the model with batch size 4, CORAL loss weight $\beta_\text{DA} = 1e8$, and the red ones denotes the model with batch size 4, CORAL loss weight $\beta_\text{DA} = 1e9$ . The localization, classification and CORAL losses are shown. The weight $\beta_\text{DA}$ was chosen to balance the contribution of the localization, classification \& CORAL losses.

\begin{table}[hb!]

\centering
\begin{tabular}{|
>{\columncolor[HTML]{F2F2F2}}c |
>{\columncolor[HTML]{FFFFFF}}c |
>{\columncolor[HTML]{FFFFFF}}c |}
\hline
{\color[HTML]{FFFFFF} \textbf{}}                                                     & \cellcolor[HTML]{0082C8}{\color[HTML]{FFFFFF} \textbf{w/o DA}} & \cellcolor[HTML]{0082C8}{\color[HTML]{FFFFFF} \textbf{with DA}} \\ \hline
{\color[HTML]{0082C8} Car-bev@0.50}                                                  & {\color[HTML]{FF0000} 8.82}                                    & {\color[HTML]{FF0000} 12.56}                                    \\ \hline
{\color[HTML]{0082C8} Car-bev@0.70}                                                  & {\color[HTML]{0082C8} 6.06}                                    & {\color[HTML]{0082C8} 4.54}                                     \\ \hline
{\color[HTML]{0082C8} Car-aos}                                                       & {\color[HTML]{FF0000} 36.77}                                   & {\color[HTML]{FF0000} 56.76}                                    \\ \hline
{\color[HTML]{0082C8} \begin{tabular}[c]{@{}c@{}}Pedestrian\\ bev@0.25\end{tabular}} & {\color[HTML]{0082C8} 0.06}                                    & {\color[HTML]{0082C8} 0.03}                                     \\ \hline
{\color[HTML]{0082C8} \begin{tabular}[c]{@{}c@{}}Pedestrian\\ bev@0.50\end{tabular}} & {\color[HTML]{0082C8} 0.00}                                    & {\color[HTML]{0082C8} 0.00}                                     \\ \hline
{\color[HTML]{0082C8} \begin{tabular}[c]{@{}c@{}}Pedestrian\\ aos\end{tabular}}      & {\color[HTML]{FF0000} 10.88}                                   & {\color[HTML]{FF0000} 12.40}                                    \\ \hline
\end{tabular}
\caption{The table demonstrates the gain in performance in 3D detection metrics, with and without the application of DA using the CORAL loss term. }
\label{tab:sim2real}
\end{table}

From these loss curves we find that the CORAL loss converges quickly in the first
several training epochs. The final converged loss shares a similar value despite the large
weighting, while using a larger batch size seems to result in a smaller CORAL loss magnitude. A
higher CORAL loss weight hinders the descent of classification/location loss curves over
the source domain simulated Shuttle A. We conclude that applying appropriate weighted CORAL loss with short training epochs improves the performance on real data. 

\section{Sources of S2R domain gap}

In our experiments with simulator we noticed a few key issues with the way simulated pointclouds were generated, and subsequently causing problems during training. We highlightg these qualitative issues for future improvement of Simulation to Real (S2R) domain transferability.

\textbf{Synthetic vs Real pointclouds} : We observed a sharp dropout noise in the real world pointclouds, while simulated pointclouds where spatio-temporally consistent. This corresponded to 30\% drop in points in the real world pointclouds. This drop in point density has a direct negative effect on the quality recall in the real world. Pointcloud subsampling procedures are a key component to tune the S2R domain gap here. This sampling pattern has been studied separately as a topic by authors in \cite{yang20203dssd}. Further on, while real world pointclouds had multiple return outputs (range) simulated pointclouds had a single return value. We also have not modeled the presence of dropout noise in the LiDAR pointclouds yet in this study, which reduces the number of points drastically. Finally real world pointclouds undergo motion distortion due to ego-vehicle motion, while simulated pointclouds do not explicitly model this behavior. This might be important for vehicles driving at high speeds.

\begin{figure*}[ht]
    \centering
    \includegraphics[width=0.33\linewidth]{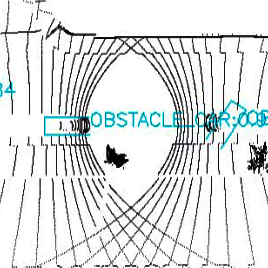}
    \hspace{2cm}
    \includegraphics[width=0.33\linewidth]{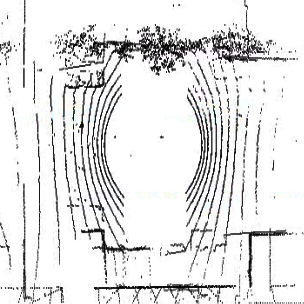}
    \caption{Simulated vs Real vehicle shadow.}
    \label{fig:sim_real_shadow}
\end{figure*}

The pointclouds in the real world contain shadowed regions with no points, created due to the presence of the ego-vehicle.  While in the simulated pointclouds, the pointclouds contain no ego-vehicle shadow. It is as if the ego-vehicle is transparent and the sensors on the vehicle gather data without any visibility through ego-vehicle taken into account. This is demonstrated via a BEV image over simulated and real vehicle pointclouds in figure \ref{fig:sim_real_shadow}. Further more objects found partially within the shadow are annotated in simulation while real pointclouds frequently contain very few points on the object. This might lead to over-fitting issues while training on simulation data.

\textbf{Pointcloud sparsity} : To avoid the sparse generated pointclouds it could be better to have a pre-processing step that removes ground truth (GT) boxes with very low point occupancy. Also, ground truth bounding boxes contain variable number of points. Thus varying the sparsity of the pointclouds in the voxels could help train a better Pointpillar backbone.

\textbf{Selecting GT boxes within sensor FOV and range} : The CARLA simulator provides bounding boxes of all agents in the virtual city irrespective of their visibility to the LiDAR sensor. Points on objects at large range values are no more visible though their corresponding bounding boxes have are still provided by CARLA. The same is applicable with objects in non line positions, eg. behind another vehicle. We manually filter out such detections by thresholding boxes based on the number of points within them.

\textbf{Simulator rendering issues} : LiDAR pointclouds are a spatio-temporal stream in the real-world while the simulated pointclouds are received as frames. This might lead to some differences in how object bounding boxes might be rendered by the simulator leading to un-correlated shifts between the bounding box and rendered pointclouds. As a result, we observed part of the points that actually belongs to a  car ( 20-30\% at maximum) may fall outside its ground truth bounding box. This issue appears in frames mainly when the objects are dynamic with respect to the ego vehicle location.

\section{Conclusion}
\hspace{\parindent}The key goal of this study is to evaluate the simulation-to-real transferability of point-cloud-based 3D object detection model performance. We evaluated the performance of the Pointpillars 3D detector on KITTI and nuScenes datasets. This model was further modified to be trained on simulated datasets generated with the CARLA simulator, before predicting on real data collected with vehicles in production. The final goal was to obtain bounding box predictions on the real vehicle pointclouds and alleviate the annotators work with automation of the annotation process.

One of the main down sides in using CORAL loss is the size of the covariance matrices over real and simulated feature maps. Large matrices can not be evaluated and thus we are limited to low resolution feature maps.

In future work we aim to study the usage of domain randomization \cite{johnson2017driving} to help reduce the simulation-to-real gap by randomizing parameters of the simulator.

\section*{Acknowledgements}
We would like to thank Barthélemy Picherit \& members of simulation team for their 
support and collaboration during this project. We thank Alexandre Almin from Navya for his comments on the paper. This work is part of the Deep Learning
Segmentation (DLS) project financed by \href{https://www.ademe.fr/}{ADEME}.

{\small
\bibliographystyle{apalike}
\bibliography{example}
}
\end{document}